\documentclass[runningheads]{llncs}

\usepackage{graphicx}
\usepackage{comment}
\usepackage{amsmath,amssymb}
\usepackage{color}
\usepackage{url}
\usepackage{hyperref}
\usepackage{tikz}
\usepackage{pgfplots}
\usepackage{subcaption}
\usepackage{cite}
\usepackage[labelformat=simple]{subcaption}

\pgfplotsset{compat = newest}

\usetikzlibrary{intersections}
\usetikzlibrary{patterns}

\usepgfplotslibrary{fillbetween}

\DeclareMathOperator{\sign}{sign}
\DeclareMathOperator{\Clip}{Clip}
\DeclareMathOperator{\adv}{adv}

\begin{document}

\title{Single-Step Adversarial Training for Semantic Segmentation}

\author{Daniel Wiens\inst{1}
\and Barbara Hammer\inst{2}}

\authorrunning{D. Wiens and B. Hammer}

\institute{Mercedes-Benz AG, 70565 Stuttgart, Germany\\ \email{daniel.wiens@daimler.com} \and Bielefeld University, 33619 Bielefeld, Germany \email{bhammer@techfak.uni-bielefeld.de}}

\maketitle

\begin{abstract}
Even though deep neural networks succeed on many different tasks including semantic segmentation, they lack on robustness against adversarial examples. To counteract this exploit, often adversarial training is used. However, it is known that adversarial training with weak adversarial attacks (e.g. using the Fast Gradient Method) does not improve the robustness against stronger attacks. Recent research shows that it is possible to increase the robustness of such single-step methods by choosing an appropriate step size during the training. Finding such a step size, without increasing the computational effort of single-step adversarial training, is still an open challenge. In this work we address the computationally particularly demanding task of semantic segmentation and propose a new step size control algorithm that increases the robustness of single-step adversarial training. The proposed algorithm does not increase the computational effort of single-step adversarial training considerably and also simplifies training, because it is free of meta-parameter. We show that the robustness of our approach can compete with multi-step adversarial training on two popular benchmarks for semantic segmentation.

\end{abstract}

\section{Introduction}
\begin{figure}[bt]
    \begin{subfigure}{.45\textwidth}
        \centering
        \includegraphics[width=1\linewidth]{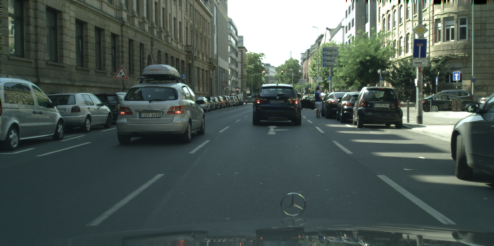}
        \caption{Clean image from Cityscapes~\cite{Cordts:2016}.}
        \label{fig:clean_input}
    \end{subfigure}\hfill
    \begin{subfigure}{.45\textwidth}
        \centering
        \includegraphics[width=1\linewidth]{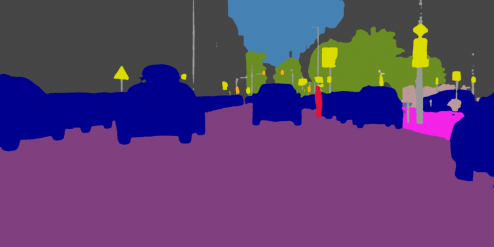}
        \caption{Prediction of the clean image.}
        \label{fig:clean_output}
    \end{subfigure}\\
    \begin{subfigure}{.45\textwidth}
        \centering
        \includegraphics[width=1\linewidth]{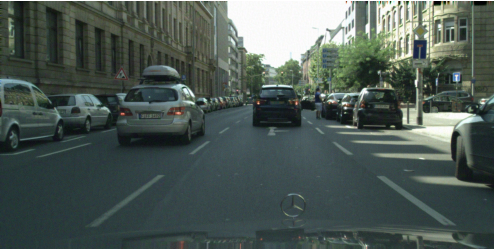}
        \caption{Adversarial example created with the Basic Iterative Method $(\varepsilon=0.03)$.}
        \label{fig:adversary_input}
    \end{subfigure}\hfill
    \begin{subfigure}{.45\textwidth}
        \centering
        \includegraphics[width=1\linewidth]{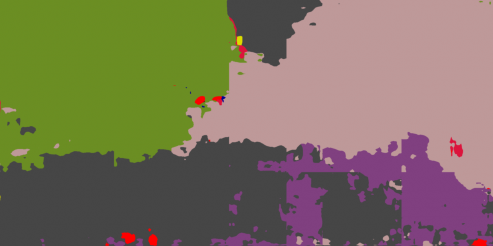}
         \caption{Prediction of the adversarial example.}
        \label{fig:adversary_output}
    \end{subfigure}
\caption{Predictions for a clean input and an adversarial example produced by a deep neural network for semantic segmentation. The two inputs in~\subref{fig:clean_input} and~\subref{fig:adversary_input} look the same for a human observer, but the predictions of the clean image and the adversarial example shown in~\subref{fig:clean_output} and in~\subref{fig:adversary_output}, respectively, are completely different.}
\label{fig:adversarialexample}
\end{figure}

Due to their great performance, deep neural networks are increasingly used on many classification tasks. Especially in vision tasks, like image classification or semantic segmentation, deep neural networks have become the standard method. However, it is known that deep neural networks are easily fooled by adversarial examples~\cite{Szegedy:2014}, i.e. very small perturbations added to an image such that neural networks classify the resulting image incorrectly. Interestingly, adversarial examples can be generated for multiple machine learning tasks, including image classification and semantic segmentation, and the perturbations are most of the time so small that humans do not even notice the changes (e.g. see fig.~\ref{fig:adversarialexample}). This phenomenon highlights a significant discrepancy of the human vision system and deep neural networks, and highlights a possibly crucial vulnerability of the latter. This fact should be taken into consideration, especially in safety critical applications like autonomous driving cars~\cite{Willers:2020}.

To increase the robustness of deep neural networks, progress along two different lines of research could be observed in the last years: provable robustness and adversarial training. Provable robustness has the goal to certify that the prediction does not change in a local surrounding for most inputs. This approach has the advantage of yielding robustness guarantees, but it is not that scalable to complex deep neural networks yet~\cite{Wong:2018,Raghunathan:2018}, or it severely affects the inference time~\cite{Cohen:2019} which is problematic for many applications. In contrast to this theoretical viewpoint, the idea of adversarial training is more empirically driven: create adversarial examples during training and use them as training data~\cite{Goodfellow:2015, Madry:2018}, this procedure can be interpreted as an efficient realization of a robustified loss function which minimizes the error simultaneously for potentially disturbed data~\cite{Shaham:2016}. Adversarial training has the advantage that it is universally applicable, and often results in high robustness albeit this holds empirically and w.r.t. a specific norm. Only reject options have the potential to improve the robustness w.r.t. different norms~\cite{Stutz:2020}.

An optimization of the inner loop of the adversarial loss function is often addressed by numeric methods which rely on an iterative perturbation of the input in the direction of its respecting gradient. But this leads to multiple forward and backward passes through the deep neural network and therefore highly increases the training time. To reduce the computational effort of adversarial training Goodfellow et al. used the information of just a single gradient for computing the adversarial examples~\cite{Goodfellow:2015}. But this kind of single-step adversarial training can be too weak and it has been observed that overfitting can take place, i.e. the specific adversarial is classified correctly, but not its immediate environment. In particular, it is not robust against multi-step attacks~\cite{Madry:2018}. Recent research discovered that this overfitting is caused by using a static step size while creating the adversarial examples for adversarial training~\cite{Kim:2020}. To overcome this, the authors of~\cite{Kim:2020} propose a method to find an ideal step size by evaluating equidistant points in the direction of the gradient. But this algorithm is in worst case as expensive as multi-step adversarial training. 

Differently to most of the previous research on adversarial robustness, we will focus on adversarial training for semantic segmentation in this work. The task of semantic segmentation is to assign every pixel of the input image a corresponding class. Because semantic segmentation needs to address localization and semantic simultaneously, it is a more complicated task than image classification~\cite{Long:2015}. Consequently, the models for semantic segmentation are generally complex, and therefore the computational effort to train such models is in principal very high. Since adversarial training also negatively affects the training time, our goal is to find a computationally efficient method which at the same time increases the robustness of semantic segmentation models effectively. 

For this purpose, we extend the idea of robust single-step adversarial training. On the one hand, we investigate single-step adversarial training for semantic segmentation as a relevant and challenging application problem for deep learning, and we demonstrate that it shares the problem of overfitting and limited robustness towards multi-step attacks with standard classification problems. On the other hand, we demonstrate that a careful selection of the step size can mitigate this problem insofar as even random step sizes improve robustness. We propose an efficient parameterless method to choose an optimum step size, which yields robustness results which are comparable to multi-step adversarial training for two popular benchmarks from the domain of image segmentation, while sharing the efficiency of single-step approaches.

\section{Background and Related Work}

\subsection{Adversarial Training}
Adversarial training for a classical image classification task tries to solve the following robust optimization problem~\cite{Shaham:2016}: given a paired sample $(x,y)$ consisting of an input sample $x \in X \subset [0,1]^{H\times W \times C}$ and an associated label $y \in Y \subset \{1, \ldots, M\}$, where $H$, $W$, $C$ and $M$ are the height, the width, the number of color channels and the number of classes, respectively, let $\ell$ denote the loss function of a deep neural network $f_{\theta}: X \to [0,1]^M$, robust learning aims for the weights $\theta$ with
\begin{align}
    \min_\theta \mathbb{E}_{(x,y) \in (X,Y)} \left(\max_{\delta \in \mathcal{B}(0, \varepsilon)} \ell(f_\theta(x+\delta),y)\right),  \label{eq:nonconvexopt}
\end{align}
where $\mathbb{E}$ and $\mathcal{B}(0,\varepsilon)$ denote the expected value and a sphere at origin with radius $\varepsilon$, respectively. The sphere is dependent on a chosen distance metric. In the context of adversarial examples most often $\mathcal{L}_0$, $\mathcal{L}_2$ and $\mathcal{L}_\infty$ are considered. In this paper we focus only on $\mathcal{L}_\infty$.

Because the loss function is highly nonlinear and non-convex even solving the inner maximization in eq.~\eqref{eq:nonconvexopt} is considered intractable~\cite{Katz:2017, Weng:2018}. Therefore, the maximization problem is often approximated by crafting adversarial examples, such that the optimization of adversarial training changes to   
\begin{align}
    \min_\theta \mathbb{E}_{(x_{\adv},y) \in (X_{\adv},Y)} ~ \ell(f_\theta(x_{\adv}),y), \label{eq:advtraining}
\end{align}
where the adversarial examples $x_{\adv} \in X_{\adv} \subset [0,1]^{H\times W \times C}$ are generated by a chosen adversarial attack, such that $\|x-x_{\adv}\|_\infty\leq \varepsilon$ holds. To guarantee the classification accuracy on the original clean samples usually $x_{\adv}$ is randomly set to be a clean sample or an adversarial example~\cite{Goodfellow:2015}. For crafting adversarial examples while training, most often gradient based methods, like the Fast Gradient Sign Method or the Basic Iterative Method, are used.  

\subsubsection{Fast Gradient Sign Method} The Fast Gradient Sign Method (FGSM) is a single-step adversarial attack which is a particularly easy and computationally cheap gradient based method, because it uses just a single gradient for calculating the adversary noise~\cite{Goodfellow:2015}. If a step size $\varepsilon$ is given, an adversarial example is determined by
\begin{align}
    x_{\adv} = x + \varepsilon \cdot \sign (\nabla_x \ell(f_\theta(x),y)). \label{eq:FGSM}
\end{align}

\subsubsection{Basic Iterative Method} The Basic Iterative Method (BIM) is a multi-step generalization of the FGSM~\cite{Kurakin:2017}. Instead of evaluating just one gradient, the BIM uses $n$ gradients iteratively, to reach a stronger adversarial example. Given a maximum perturbation $\varepsilon$, a number of iterations $N$ and a step size $\alpha$, an adversarial example created with BIM is given iteratively by
\begin{equation}\label{eq:BIM}
\begin{split}
    x_0 &= x,\\
    x_{i+1} &= \Clip_{x,\varepsilon} (x_i + \alpha \cdot \sign (\nabla_x \ell(f_\theta(x_i),y))), \\
    x_{\adv} &= x_N.   
\end{split}
\end{equation}
Where the $\tilde{p}=\Clip_{x,\varepsilon}(p)$-Function projects $p$ into the $\varepsilon$-neighbourhood of x, such that $\|\tilde{p}-x\|_\infty \leq \varepsilon$.

\subsection{Restrictions of Single-step Adversarial Training}
To generate adversarial examples while training, Goodfellow et al. chose the FGSM as a computationally cheap adversarial attack~\cite{Goodfellow:2015}. It was shown that models trained with such a single-step adversarial training are not robust against multi-step attacks~\cite{Madry:2018}. To train robust models, multi-step adversarial training is commonly used. Since multi-step approaches require multiple forward and backward passes through the deep neural network, training a robust model becomes computationally very intensive.

To minimize the computational effort while training, one line of work tries to increase the robustness of single-step adversarial training. Wong et al. analyzed the robustness of single-step adversarial training epoch-wise and observed that the robustness against BIM increases in the beginning during training, but decreases after some epochs~\cite{Wong:2020}. This observation is called catastrophic overfitting. To overcome this phenomenon Wong et al. added random noise to the input before using the FGSM for adversarial training and also added early-stopping by tracking the multi-step robustness of small batches. A few further approaches follow this line of research \cite{Li:2020, Andriushchenko:2020}.

\pgfplotscreateplotcyclelist{MyList_sketch}{
thin,black,solid\\
thin,red,dashed\\
}
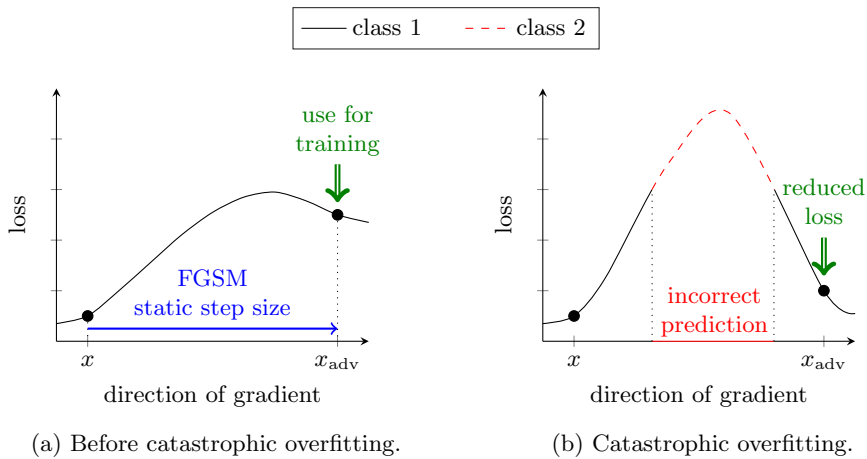
\begin{figure}[bt]
     \begin{subfigure}{1\textwidth} 
    \begin{tikzpicture}
        \begin{axis}[
          cycle list name=MyList_sketch,
          width=1\textwidth, height=0.3\textwidth,
          hide axis,
          xmin=0, xmax=10, ymin=0, ymax=0.2,
          legend columns=3,
          legend style={ 
          /tikz/every even column/.append style={column sep=0.5cm},
             draw=white!15!black,
             legend cell align=left,
              at={(0.75,0.5)}
             }]

            \pgfplotsinvokeforeach{1,2}{\addplot coordinates {(0,-1)};}
            
            \addlegendentry{class 1}
            \addlegendentry{class 2}
        \end{axis}
    \end{tikzpicture}
    \end{subfigure}\hfill

    \begin{subfigure}{.47\textwidth}
    \begin{tikzpicture}
        \begin{axis}[
                    width=1\textwidth,
                    ymin=0, 
                    ymax=1,
                    ylabel={loss},
                    xmin=0.,
                    xmax=1.,
                    xtick={0.1,0.9},
                    xticklabels = {$x$, $x_{\adv}$},
                    xlabel={direction of gradient},
                    axis lines=left,
                    ytick={0.2,0.4,0.6,0.8},
                    yticklabels = {,,,},
                    clip = false,
                    ]
            \addplot[draw=black, smooth, name path=f] plot coordinates
                {
                (0,0.07)
                (0.1, 0.1)
                (0.2,0.2)
                (0.3,0.31)
                (0.4,0.42)
                (0.5, 0.51)
                (0.6,0.57)
                (0.7,0.59)
                (0.8,0.555)
                (0.9, 0.5)
                (1, 0.47)
                };
            \addplot[dotted] plot coordinates
                {
                (0.1, 0.1)
                (0.1,0)
                };
            \addplot[dotted] plot coordinates
                {
                (0.9, 0.5)
                (0.9,0)
                };
            \addplot[->, blue, thick] plot coordinates
                {
                (0.1, 0.05)
                (0.9,0.05)
                };
            \node[blue,above,align=center] at (axis cs:0.5,0.05){\small{FGSM} \\ \small{static step size}};
            \addplot[only marks, mark options={solid,draw=black,fill=black}]plot coordinates 
                {
                (0.1, 0.1)
                (0.9, 0.5)
                 };
            \addplot[->,double, green!50!black, thick] plot coordinates
                {
                (0.9, 0.7)
                (0.9,0.55)
                };
            
            \node[green!50!black,above,align=center] at (axis cs:0.9,0.7){\small{use for}\\ \small{training}};
        \end{axis}
    \end{tikzpicture}
    \caption{Before catastrophic overfitting.}
    \label{fig:beforecatastrophicoverfitting}
    \end{subfigure}\hfill
    \begin{subfigure}{.47\textwidth}
    \begin{tikzpicture}
        \begin{axis}[
                    width=1\textwidth,
                    ymin=0, 
                    ymax=1,
                    ylabel={loss},
                    xmin=0.,
                    xmax=1.,
                    xtick={0.1,0.9},
                    xticklabels = {$x$, $x_{\adv}$},
                    xlabel={direction of gradient},
                    ytick={0.2,0.4,0.6,0.8},
                    yticklabels = {,,,},
                    axis lines=left,
                    clip = false,
                    ]
            \addplot[draw=none, thick, smooth, name path=g] plot coordinates
                {
                (0,0.07)
                (0.1, 0.1)
                (0.2, 0.25)
                (0.4, 0.7)
                (0.6, 0.9)
                (0.9, 0.2)
                (1,0.11)
                };
            \addplot[draw=none, name path=g2] plot coordinates
                {
                (0,0.6)
                (1,0.6)
                };
            \path [
                draw,
                color=black,
                intersection segments={
                    of=g and g2 ,
                    sequence={A0}
                }];
            \path [
                draw,
                color=black,
                intersection segments={
                    of=g and g2 ,
                    sequence={A2}
                }];
            \addlegendimage{line width=0.3mm,color=black}
            \path [
                draw,
                color=red,
                dashed,
                intersection segments={
                    of=g and g2 ,
                    sequence={A1}
                }]; 
            \addlegendimage{line width=0.3mm,color=green}
            \addplot[only marks, mark options={solid,draw=black,fill=black}]plot coordinates 
                {
                (0.1, 0.1)
                (0.9, 0.2)
                 };
             \node[red,below,align=center] at (axis cs:0.55,0.25){\small{incorrect}\\ \small{prediction}};
            \node[green!50!black,above,align=center] at (axis cs:0.9,0.43){\small{reduced}\\ \small{loss}};
            \addplot[->,double, green!50!black, thick] plot coordinates
                {
                (0.9, 0.41)
                (0.9,0.26)
                };
                
            \addplot[draw=red, name path=B] plot coordinates
                {
                (0.35,0)
                (0.74,0)
                };    
            \addplot[color=gray!0,draw=none]
            fill between[
                of=g and B,
                soft clip={domain=0.35:0.74},
            ];
            \addplot[dotted] plot coordinates
                        {
                        (0.35, 0.)
                        (0.35,0.6)
                        };
            \addplot[dotted] plot coordinates
                {
                (0.74, 0.)
                (0.74,0.6)
                };
        \end{axis}
    \end{tikzpicture}
    \caption{Catastrophic overfitting.}
    \label{fig:aftercatastrophicoverfitting}
    \end{subfigure}
    \caption{Sketch: The static step size of single-step adversarial training as reason for catastrophic overfitting. \subref{fig:beforecatastrophicoverfitting}~After initial single-step adversarial training the adversarial example is classified correctly. \subref{fig:aftercatastrophicoverfitting}~Further training reduces the loss of the adversarial example in fixed distance to the original sample even more, but in between the loss increases resulting in a region of incorrect predictions.}
    \label{fig:RadicalOverfitting}
\end{figure}
Because these methods need to calculate more than one gradient at some point, they are computationally inefficient. Kim et al. showed empirically that the static step size for generating the adversarial examples constitute a reason why single-step adversarial training is not robust~\cite{Kim:2020}. Single-step adversarial examples which are generated by a fixed step size have a fixed distance from their original data points. When training on such samples as well as on the original data points, there is no incentive of the loss function to predict the interval in between these extremes correctly or to provide a monotonic output in this region. This can lead to catastrophic overfitting as displayed in fig.~\ref{fig:RadicalOverfitting}, i.e. the original data and the chosen adversarial examples are correct, but not the region in between. Multi-step adversarial attacks iteratively use small step sizes, and therefore are able to find these areas of non-monotonicity, such that single-step adversarial training is not robust against multi-step attacks. To overcome this problem, Kim et al. proposed to test a range of different step sizes by equidistant sampling in the direction of the gradient and to choose the smallest step size which leads to a false prediction for generating the adversarial example~\cite{Kim:2020}. Since this approach aims for a minimum step size to generate an adversarial sample, we can expect monotonicity of the network output in between, at least it is guaranteed that no closer adversarials can be found in the direction of the gradient. Hence overfitting seems less likely in such cases, as is confirmed by the experimental finding presented in~\cite{Kim:2020}.

Yet, the sampling strategy which is presented in~\cite{Kim:2020} requires several forward passes through the network and is thus in worst case as demanding as multi-step approaches. In the following, we propose an approximate analytic solution how to compute a closest adversarial, which leads to an efficient implementation of this robust step size selection strategy.

\section{Methods}
\subsection{Choosing the Step Size for Image Classification}\label{sec:imageclassification}
To explain the idea behind our step size control algorithm, we first start looking at image classification. We are interested in an efficient analytic approximation which yields the closest adversarial example, i.e. a closest pattern where the classification changes. The deep neural network is given as a function $f_\theta:X \to [0,1]^M$, where the output of the deep neural network $f_\theta(x)$ is computed using the softmax function. The inputs to the softmax $z(x)$ are called logits. We define the gain function $g:X \times Y \to \mathbb{R}$ as 
\begin{equation}
 g(x,y) = z_y(x)-\max_{i \neq y} z_i(x), \label{eq:ourgainfunc}
\end{equation}
where $z_i(x)$ is the logit value of class $i$. The gain function in eq.~\eqref{eq:ourgainfunc} has the property that
\begin{equation}
    g(x,y) \left\{\begin{array}{ll} >0 & \quad \text{if }\arg \max f_\theta(x) = y\\
         \leq 0  & \quad \text{else}\end{array}\right. . \label{eq:gainproperty}
\end{equation}
For given $(x,y)$ a close adversarial example can be found at the boundary between a correct and an incorrect prediction, which holds if  
\begin{equation}
    g(x_{\adv},y) = 0. \label{eq:setgaintozero}
\end{equation}
Using the update rule of FGSM in eq.~\eqref{eq:FGSM} on the gain function, we create an adversarial example by
\begin{equation}
    x_{\adv} = x - \varepsilon \cdot \sign( \nabla_x g(x,y)). \label{eq:FGSMongain}
\end{equation}
For estimating the step size $\varepsilon$, we linearly approximate the gain function $g$ with a Taylor approximation
\begin{equation}
    g(x_{\adv},y) \approx g(x,y)+(x_{\adv}-x)^T \cdot \nabla_x g(x,y). \label{eq:taylor}
\end{equation}
Combining eq.~\eqref{eq:setgaintozero}, eq.~\eqref{eq:FGSMongain} and eq.~\eqref{eq:taylor} results in 
\begin{equation}
    0 = g(x,y)- \varepsilon \cdot \sign( \nabla_x g(x,y))^T \cdot \nabla_x g(x,y),
\end{equation}
which we can solve for the step size $\varepsilon$:
\begin{align}
\varepsilon = \frac{g(x,y)}{\nabla_x g(x,y)^T \cdot \sign(\nabla_x g(x,y))} = \frac{g(x,y)}{\|\nabla_x g(x,y)\|_1}.
\end{align}
Notice, the idea of the above step size control is the same as doing one step of Newton's approximation method for finding the zero crossings of a function. We therefore call our approach Fast Newton Method. 
Contrary to the classical solution of robust optimization, we use an approximation of the closest adversarial for training. This has two effects: on the one hand, we expect monotonicity of the loss in between, or a sufficient distance from $0$. Hence a network which classifies $x$ and the closest adversarial correctly likely classifies the enclosed interval in the same way, avoiding catastrophic overfitting. On the other hand, we do not consider any predefined radius; rather we use the smallest radius that includes an adversarial example. Using these adversarial examples for training, iteratively increases the radius of the sphere, and consequently the distance where adversarials can be found.

\subsection{Robust Semantic Segmentation}
The above mentioned results were presented in the context of image classification. In this work we want to concentrate on semantic segmentation. It is already known that deep neural networks for semantic segmentation are also vulnerable against adversarial examples~\cite{Xie:2017, Metzen:2017}, but only few approaches address the question how to efficiently implement robust training for image segmentation tasks. As far as we know, there is only one work investigating the impact of multi-step adversarial training on semantic segmentation~\cite{Xu:2020}.

First, we formalize the learning objective for robust semantic segmentation. Because semantic segmentation determines the class of each single pixel of an image, semantic segmentation could be interpreted as multiple pixel-wise classifications. To predict all the pixels at the same time the deep neural network function $f_\theta:X \to [0,1]^{H \times W \times M}$ has increased output dimension. The set of labels are given by $Y \subset \{1, \ldots, M\}^{H \times W}$. Robust learning on semantic segmentation tries to optimize 
\begin{align}
    \min_\theta \mathbb{E}_{(x,y) \in (X,Y)} \left(\max_{\delta \in \mathcal{B}(0, \varepsilon)} \frac{1}{H W}\sum_{j=0}^{H W}\ell(f_{\theta,j}(x+\delta),y_j)\right)  \label{eq:nonconvexoptsemsec}
\end{align}
for the weights $\theta$, where $f_{\theta,j}(x)$ and $y_j$ are the prediction and the label of the j-th pixel, respectively. To find the weights for all the pixel-wise predictions simultaneously, the losses of the pixel-wise predictions are averaged.\\
Adversarial training on semantic image segmentation approximates the min-max loss by the following term:
\begin{align}
    \min_\theta \mathbb{E}_{(x_{\adv},y) \in (X_{\adv},Y)}~ \frac{1}{H W}\sum_{j=0}^{H W}\ell(f_{\theta,j}(x_{\adv}),y_j),  \label{eq:robustoptsemantic}
\end{align}
where $x_{\adv} \in X_{\adv}$ constitutes a suitable image with $\|x-x_{\adv}\|_\infty\leq \varepsilon$, which plays the role of an adversarial in the sense that it leads to an error of the segmented image for a large number of pixels. Yet, unlike for scalar outputs, it is not clear what exactly should be referred to by an adversarial: we can aim for an input $x_{\adv}$ such that all output pixels change, or, alternatively, approximate this computationally extensive extreme by an efficient surrogate, as we will introduce in the following.

\subsection{Choosing the Step Size for Semantic Segmentation}\label{sec:stepsizesemseg}
Initially, we treat each pixel as a separate output. An adversarial corresponds to an input such that one specific output pixel changes. For this setting, the gain function from eq.~\eqref{eq:ourgainfunc} becomes: 
\begin{equation}
    g_j(x, y_j) = z_{j,y_j}(x)- \max_{i \neq y_j} z_{j,i}(x), \quad \text{for } j \in \{1,\ldots,H W\},
\end{equation}
where $z_{j,i}(x)$ is the logit value of the $j$-th pixel of class $i$. To find a close adversarial example $x_{\adv}$ which changes all (or a large number of) pixels, each pixel-wise output should be at the boundary between a correct and an incorrect prediction. This holds if
\begin{equation}
    g_j(x_{\adv},y_j)=0, \quad \forall j \in \{1,\ldots,H W\}.
\end{equation}
Using our results from sec.~\ref{sec:imageclassification}, we obtain a different step size $\varepsilon_j$ for every pixel, hence a possibly different adversarial $x_{\adv}^{(j)}$ for each pixel-wise prediction. A common adversarial example $x_{\adv}$ which approximately suits for all pixel-wise predictions in the same time, could be chosen as the average term
\begin{equation}
    x_{\adv}=\frac{1}{H W}\sum_{j=1}^{H W} x_{\adv}^{(j)}.
\end{equation}
But this procedure requires the computation of $H W$ different gradients $\nabla_x g_j(x, y_j)$ for $j=1,\ldots,H W$ which is computational too expensive to work with.

As an alternative, we can refer to adversarial examples as images for which a number of pixels change the segmentation assignment. In this case we consider the pixel-wise predictions at the same time from the beginning by using the average of the gain functions (rather than the average of the specific pixel-wise adversarials). A close adversarial example in this sense is then given if
\begin{equation}
    \frac{1}{H W}\sum_{j=1}^{H W} g_j(x_{\adv},y_j)=0.
\end{equation}
Defining the averaged gain function as $\bar{g}(x,y) = \frac{1}{H W}\sum_{j=1}^{H W} g_j(x,y_j)$ and using our method from sec.~\ref{sec:imageclassification} results in a single step size
\begin{align}
\varepsilon = \frac{\bar{g}(x,y)}{\|\nabla_x \bar{g}(x,y)\|_1}, \label{eq:finaleps}
\end{align}
such that the update rule for finding an adversarial example becomes
\begin{align}
    x_{\adv} = x -  \frac{\bar{g}(x,y)}{\|\nabla_x \bar{g}(x,y)\|_1} \cdot \sign( \nabla_x \bar{g}(x,y)).
\end{align}
Of course using the averaged gain function for calculating the step size $\varepsilon$ in eq.~\eqref{eq:finaleps} leads to a more loose approximation, but as a trade off this method does not increase the computational effort compared to adversarial training with the FGSM considerably, because we also calculate just one gradient.

\section{Experiments}
\subsection{Implementation} 
\subsubsection{Datasets} To evaluate our approach, we use the datasets Cityscapes~\cite{Cordts:2016} and PASCAL VOC~\cite{Everingham:2015}. Cityscapes contains $2975$, $500$ and $1525$ colored images of size $1024 \times 2048$ for training, validation and testing, respectively. The labels assign most pixels one of $19$ classes, where some areas are not labeled.\\
PASCAL VOC originally includes $1464$, $1499$ and $1456$ differently sized images for training, validation and testing, respectively. Later the training set was increased to $10582$ images~\cite{Hariharan:2011}. Including the background class, the pixels are labeled as one of $21$ classes. Like in Cityscapes there are also some areas not labeled, such that we limit our loss and gain function for both datasets to the labeled pixels. We always normalise the color channels in the range of $[0,1]$.

\subsubsection{Model} As base model we chose the popular PSPNet50 architecture from~\cite{Zhao:2017} with a slight modification. Instead of using ResNet50 we work with the improved ResNet50v2 as fundamental pretrained network~\cite{He:2016}. For the training parameters we follow the paper~\cite{Zhao:2017} using SGD with a momentum of $0.9$. The learning rate decays polynomially with a base learn rate of $0.01$ and power $0.9$. We use a batchsize of $16$ and also include $\mathcal{L}_2$ weight decay of $10^{-4}$. However, for a better comparison to different models, we do not apply the auxiliary loss from~\cite{Zhao:2017}. To further increase the variation of the dataset, the data is randomly horizontally flipped, resized between $0.5$ and $2$, rotated between $-10$ and $10$ degrees, and additionally randomly blurred. Afterwards, we randomly crop the images to a size of $712 \times 712$ and $472 \times 472$ for Cityscapes and PASCAL VOC, respectively. For evaluation, we will compare the robustness of the following models:
\begin{itemize}
    \item The Basic Model without using any adversarial examples in training,
    \item The FGSM Model trained with FGSM at $\varepsilon =0.03$,
    \item The random FGSM Model trained with FGSM at $\varepsilon \sim \mathcal{U}(0,0.03)$ chosen uniformly,
    \item The BIM Model trained with BIM ($\alpha=0.01$, $\varepsilon =0.03$, $N=3$),
    \item Fast Newton Method trained with our algorithm,
\end{itemize}
where the parameters of the BIM Model are taken from~\cite{Xu:2020}. For adversarial training the input is randomly chosen either a clean data sample or an adversarial example, each with probability of 0.5.

\subsubsection{Evaluating the Robustness} We measure the performance on the clean and the adversary data with mean IoU~\cite{Everingham:2015}. The mean IoU is a standard measure for semantic segmentation. It is used over the standard accuracy per Pixel, because it better represents the desired accuracy in case of differently sized objects.

For the empirical robustness evaluation we attack each model with FGSM and BIM. For both attacks we vary the attack radius $\varepsilon$ from $0$ to $0.04$, such that we receive one robustness curve for each model and attack. As trade off between computational time and attack strength, we use BIM with $N=10$ iterations. We chose the step size $\alpha$ as small as possible, such that the maximum perturbation rate of $\varepsilon=0.04$ is still reachable. Hence, we use $\alpha=0.004$ as step size for BIM.

\subsection{Results on Cityscapes}

\pgfplotscreateplotcyclelist{MyList}{
thin,blue,solid,mark=triangle*\\
thin,brown,solid,mark=star\\
thin,green,solid,mark=diamond*\\
thin,solid,mark=square*,color=black\\
thin,solid,mark=*,color=red\\
}

\begin{figure}[bt]
    \begin{subfigure}{1\textwidth} 
    \begin{tikzpicture} 
        \begin{axis}[
          cycle list name=MyList, 
          width=1\textwidth, height=0.32\textwidth,
          hide axis,
          xmin=0, xmax=10, ymin=0, ymax=0.2,
          legend columns=3,
          legend style={ 
          /tikz/every even column/.append style={column sep=0.89cm},
             draw=white!30!black,
             legend cell align=left,
              at={(1.,0.5)},
              font = \small
             }]

            \pgfplotsinvokeforeach{1,2,3,4,5}{\addplot coordinates {(0,-1)};}
            \addlegendentry{Basic Model}
            \addlegendentry{FGSM Model}
            \addlegendentry{random FSGM Model}
            \addlegendentry{BIM Model}
            \addlegendentry{Fast Newton Method}
        \end{axis}
    \end{tikzpicture}
    \end{subfigure}\hfill\\
    \pgfplotstableread{Data/FGSM_results.csv}{\table}
    \begin{subfigure}{.45\textwidth}
        \begin{tikzpicture}
            \begin{axis}[
                    cycle list name=MyList,
                    ymin=0, 
                    ymax=0.8,
                    xmin=0.,
                    xmax=0.04,
                    scaled ticks=false,
                    tick label style={/pgf/number format/fixed, font=\scriptsize},
                    xlabel={$\varepsilon$},
                    ylabel={mean IoU},
                    ylabel style = {font=\normalsize},
                    xlabel style = {font=\normalsize},
                    width=1\textwidth,
                    axis lines*=left,
                    ]
                \addplot table [x = {0}, y = {1}] {\table};

                \addplot table [x = {0}, y = {2}] {\table};
                    
                 \addplot table [x = {0}, y = {3}] {\table};
                
                 \addplot table [x = {0}, y = {4}] {\table};
                    
                 \addplot table [x = {0}, y = {5}] {\table};
            \end{axis}
        \end{tikzpicture}
        \caption{Robustness against single-step adversarial examples created with FGSM.\\ ~}
        \label{fig:FGSMcurves}
    \end{subfigure}\hfill
    \pgfplotstableread{Data/BIM_results.csv}{\table}
    \begin{subfigure}{.45\textwidth}
        \begin{tikzpicture}
            \begin{axis}[
                    cycle list name=MyList,
                    ymin=0, 
                    ymax=0.8,
                    xmin=0.,
                    xmax=0.04,
                    scaled ticks=false,
                    tick label style={/pgf/number format/fixed, font=\scriptsize},
                    xlabel={$\varepsilon$},
                    ylabel={mean IoU},
                    ylabel style = {font=\normalsize},
                    xlabel style = {font=\normalsize},
                    width=1\textwidth,
                    legend style={font=\normalsize},
                    axis lines*=left,
                    ]
                \addplot table [x = {0}, y = {1}] {\table};
                    
                \addplot table [x = {0}, y = {2}] {\table};
                    
                 \addplot table [x = {0}, y = {3}] {\table};
                
                 \addplot table [x = {0}, y = {4}] {\table};
                    
                 \addplot table [x = {0}, y = {5}] {\table};
            \end{axis}
        \end{tikzpicture}
        \caption{Robustness against multi-step adversarial examples created with BIM ($N=10$, $\alpha=0.004$).}
        \label{fig:BIMcurves}
    \end{subfigure}

    \label{fig:RobustnessCurves}
    \caption{Robustness curves on the Cityscapes dataset.}

\end{figure}
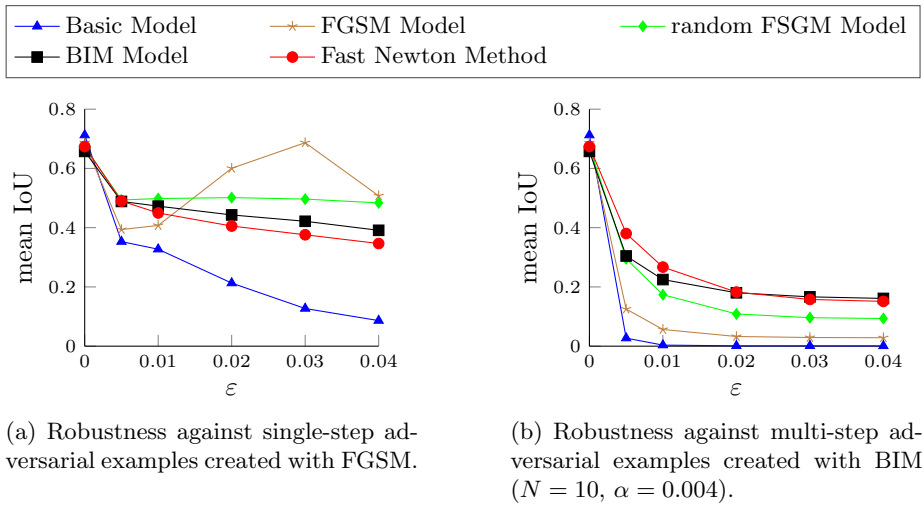

\subsubsection{Single-Step Robustness} In fig.~\ref{fig:FGSMcurves} are the summarized results regarding the FGSM shown. We see that the mean IoU of all models except the FGSM Model and the random FGSM Model decrease with increased attack strength~$\varepsilon$, whereas the performance of the BIM Model and our Fast Newton Method decrease slower. The BIM Model shows slightly better robustness than ours, but the FGSM Model and the random FGSM Model perform even better. Looking closer to the robustness curve of the FGSM Model, we observe that the model is most robust to adversarial examples with attack strength $\varepsilon=0$ and $\varepsilon=0.03$. For the other values of~$\varepsilon$ the robustness decreases. This observation matches with the phenomenon of catastrophic overfitting shown in fig.~\ref{fig:RadicalOverfitting}. The model is most robust exactly against the adversarial examples it was trained for. For the values between $\varepsilon=0$ and $\varepsilon=0.03$ the nonlinear character of deep neural networks leads to a drop in the robustness. 

\subsubsection{Multi-Step Robustness} Fig.~\ref{fig:BIMcurves} shows the robustness of the trained models against attacks with BIM. As can be seen, all models perform significantly worse against this attack. Because the robustness is defined as the performance under the worst case attack, this evaluation represents the real robustness of the models far better. We can see that the Basic Model and FGSM Model drop very fast with increased attack strength~$\varepsilon$. So we can confirm that the FGSM Model overfits. Even though the Fast Newton Method and the random FGSM Model are also single-step adversarial training models, they perform significantly better under the attack with BIM than the FGSM Model. That shows that the idea of controlling the step sizes of adversarial attacks makes single-step adversarial training more effective. The outstanding performance of our Fast Newton Method compared to the FGSM models shows the importance of choosing the correct step size for creating adversarial examples during training. Additionally, our Fast Newton Method shows for small $\varepsilon$ even better robustness than the more sophisticated BIM model, and is for larger $\varepsilon$ equally robust while being significantly less computational expensive.

\subsection{Results on VOC}
\begin{figure}[bt]
    \pgfplotstableread{Data/FGSM_results_on_VOC.csv}{\table}
    \begin{subfigure}{1\textwidth} 
    \begin{tikzpicture}
        \begin{axis}[
          cycle list name=MyList,
          width=1\textwidth, height=0.32\textwidth,
          hide axis,
          xmin=0, xmax=10, ymin=0, ymax=0.2,
          legend columns=3,
          legend style={ 
          /tikz/every even column/.append style={column sep=0.89cm},
             draw=white!30!black,
             legend cell align=left,
              at={(1.,0.5)},
              font = \small
             }]

            \pgfplotsinvokeforeach{1,2,3,4,5}{\addplot coordinates {(0,-1)};}
            \addlegendentry{Basic Model}
            \addlegendentry{FGSM Model}
            \addlegendentry{random FSGM Model}
            \addlegendentry{BIM Model}
            \addlegendentry{Fast Newton Method}
        \end{axis}
    \end{tikzpicture}
    \end{subfigure}\hfill\\
    \begin{subfigure}{.45\textwidth}
        \begin{tikzpicture}
            \begin{axis}[
                    cycle list name=MyList,
                    ymin=0, 
                    ymax=0.8,
                    xmin=0.,
                    xmax=0.04,
                    scaled ticks=false,
                    tick label style={/pgf/number format/fixed, font=\scriptsize},
                    xlabel={$\varepsilon$},
                    ylabel={mean IoU},
                    ylabel style = {font=\normalsize},
                    xlabel style = {font=\normalsize},
                    width=1\textwidth,
                    axis lines*=left,
                    ]
                \addplot table [x = {0}, y = {1}] {\table};

                \addplot table [x = {0}, y = {2}] {\table};
                    
                 \addplot table [x = {0}, y = {3}] {\table};
                
                 \addplot table [x = {0}, y = {4}] {\table};
                    
                 \addplot table [x = {0}, y = {5}] {\table};
            \end{axis}
        \end{tikzpicture}
        \caption{Robustness against adversarial examples created with FGSM.\\ ~}
        \label{fig:FGSMcurves_VOC}
    \end{subfigure}\hfill
    \pgfplotstableread{Data/BIM_results_on_VOC.csv}{\table}
    \begin{subfigure}{.45\textwidth}
        \begin{tikzpicture}
            \begin{axis}[
                    cycle list name=MyList,
                    ymin=0, 
                    ymax=0.8,
                    xmin=0.,
                    xmax=0.04,
                    scaled ticks=false,
                    tick label style={/pgf/number format/fixed, font=\scriptsize},
                    xlabel={$\varepsilon$},
                    ylabel={mean IoU},
                    ylabel style = {font=\normalsize},
                    xlabel style = {font=\normalsize},
                    width=1\textwidth,
                    legend style={font=\normalsize},
                    axis lines*=left,
                    ]
                \addplot table [x = {0}, y = {1}] {\table};
                    
                \addplot table [x = {0}, y = {2}] {\table};
                    
                 \addplot table [x = {0}, y = {3}] {\table};
                
                 \addplot table [x = {0}, y = {4}] {\table};
                    
                 \addplot table [x = {0}, y = {5}] {\table};
            \end{axis}
        \end{tikzpicture}
        \caption{Robustness against adversarial examples created with BIM ($N=10$, $\alpha=0.004$).}
        \label{fig:BIMcurves_VOC}
    \end{subfigure}
    \caption{Robustness curves on the PASCAL VOC dataset.}
    \label{fig:RobustnessCurves_VOC}
\end{figure}
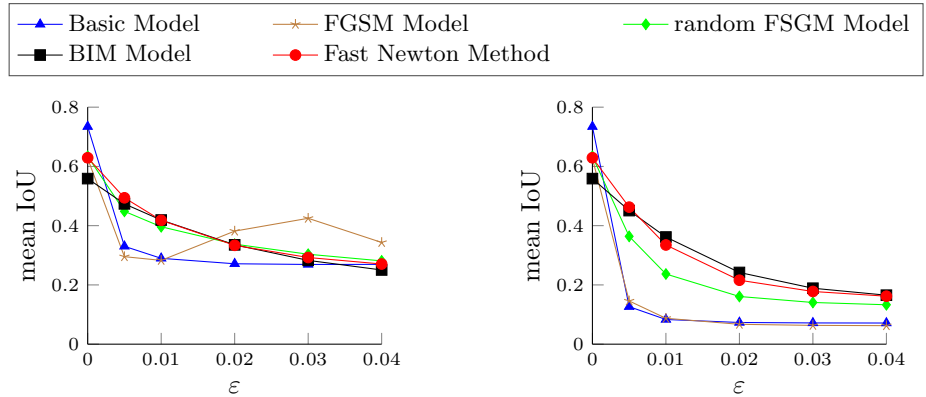

\subsubsection{Single-Step Robustness} The robustness curves against the FGSM for models trained on PASCAL VOC are shown in fig.~\ref{fig:FGSMcurves_VOC}. We observe that the attack is not as effective on the Basic Model as on Cityscapes. We think that this is mainly reasoned in the characteristic of the dataset with its dominant background class. It seems difficult for the attack to switch the class of the whole background area. But as we already mentioned, the robustness is better represented by a strong multi-step attack. Our focus here lays again in the curve course of the FGSM Model. Like on Cityscapes the FGSM Model is most robust against the adversarial examples it was trained for ($\varepsilon=0$ and $\varepsilon=0.03$). So we see again an indicator for catastrophic overfitting.

\subsubsection{Multi-Step Robustness} Looking in Fig.~\ref{fig:BIMcurves_VOC} on the robustness of the trained models against the BIM, we can observe very similar results like on Cityscapes. Both, the Basic and the FGSM Model, perform much worse than the other models. So we can clearly say that the FGSM Model overfits to adversarial examples it was trained with. However, even if the random FGSM Model and the Fast Newton Method are trained with single-step adversarial examples, too, they perform significantly better than the FGSM Model. Thus, we can confirm that varying the step size increases the robustness of such models. Because the Fast Newton Method outperforms the random FGSM Model, we conclude that our proposed step size is more superior than determining the step size randomly. Additionally, even if our Fast Newton Method is less computational expensive than the BIM Model, their robustness is very similar.

\section{Conclusion}
The research community focused so far on improving the robustness of deep neural networks for image classification. We on the other hand concentrate on semantic segmentation. We showed that single-step adversarial training for semantic segmentation underlies the same difficulties regarding the robustness against multi-step adversarial attacks. One reason for that non-robustness of single step adversarial training is the static step size for finding the adversarial examples while training. Therefore, we presented a step size control algorithm which approximates an appropriate step size for every input such that the robustness of single-step adversarial training increases significantly. As our method approximates the best step size based on the gradient, which needs to be calculated anyway for adversarial training, our method does not considerably increase the computational effort while training a significantly more robust model. In addition, our approach is easy to use, because it is free of any parameter. Finally, we showed on the datasets, called Cityscapes and PASCAL VOC, that our method equals in performance with the more sophisticated and a lot more computationally expensive multi-step adversarial training.

\end{document}